\documentclass[runningheads]{llncs}
\usepackage[T1]{fontenc}
\usepackage{amsmath,amssymb,amsfonts}
\usepackage{array}
\usepackage{adjustbox}
\usepackage{tabularx}
\usepackage{booktabs}
\usepackage{makecell}
\usepackage{multirow}
\usepackage{xcolor}
\usepackage{graphicx}
\usepackage{url}
\usepackage{todonotes}
\usepackage[inline]{enumitem}
\usepackage{algorithm}
\usepackage{algorithmicx}
\usepackage{algpseudocode}
\usepackage{listings}
\usepackage{tcolorbox}
\usepackage{arydshln}
\usepackage{pifont}
\usepackage{orcidlink}

\newcolumntype{Y}{>{\centering\arraybackslash}X}
\providecommand{\doi}[1]{\url{https://doi.org/#1}}
\begin{document}
\title{Detecting Knowledge Inconsistencies Across Text, Tables, and Knowledge Graphs}
\titlerunning{Detecting Cross-Modal Knowledge Inconsistencies}
%
%
\author{
Fanfu Wei\inst{1}\orcidID{0009-0008-4397-0492}
\thanks{Corresponding author.}
\and
Thibault Ehrhart\inst{1}\orcidID{0000-0003-1377-8279}
\and
Raphaël Troncy\inst{1}\orcidID{0000-0003-0457-1436}
}

\authorrunning{F. Wei et al.}
\institute{EURECOM, France\\
\email{\{fanfu.wei, thibault.ehrhart, raphael.troncy\}@eurecom.fr}}
\maketitle              
\begin{abstract}
Wikipedia and Wikidata are widely used for information access, LLM pre-training, and retrieval-augmented generation. Their knowledge is deeply connected but scattered across text, tables, and knowledge graphs. This raises a practical question: when these modalities disagree, how can we detect and explain the conflict? We study this problem as \emph{modality-level inconsistency detection}. We first introduce a taxonomy of cross-modal knowledge inconsistencies, covering information granularity differences, direct conflicts, temporal changes, and KG incompleteness. We then present \textsc{Kontrast}, an automatic framework that uses Text-to-SPARQL and LLM reasoning to compare table-based answers with KG evidence and categorize the resulting inconsistencies. Experiments on various Table-QA datasets show that cross-modal inconsistencies are common and informative. They reveal not only true knowledge conflicts, but also missing KG structure and temporal mismatches while being limited by Text-to-SPARQL errors and noise. Our analysis shows that text, tables, and KGs can complement and correct one another through systematic comparison. \textsc{Kontrast} provides a practical tool for large-scale knowledge auditing and establishes a benchmark for future work on cross-modal knowledge consistency. Code and data are available at \url{https://github.com/ECLADATTA/KONTRAST}.

\end{abstract}

\section{Introduction}
\label{sec:introduction}
Wikipedia and Wikidata are two of the most widely used open knowledge resources on the Web. They support human information access and serve as important knowledge sources for open-domain question answering~\cite{karpukhin-etal-2020-dense}, retrieval-augmented generation~\cite{Lewis2020}, large language model pre-training~\cite{Brown2020,Gao2020}, and other knowledge-intensive applications. Their value comes from scale, openness, and constant updates. These same properties, however, make consistency difficult to maintain.

This difficulty is amplified by the way knowledge is represented. The same fact may appear in a sentence, an infobox, or a table row of a Wikipedia article, and in a Wikidata statement. These representations are connected, but they are edited under different schemas, workflows, and update cycles. As a result, the same question can receive different answers depending on whether we consult text, tables, or the KG. A Wikipedia table may list the complete casting of a movie while Wikidata contains only the lead actors; Wikidata may store country-specific release dates for the movie while a table reports only a single year; an elevation value may differ between an infobox and Wikidata because the sources refer to different measurement points. Such cases are not merely retrieval errors, they reveal \emph{modality-level knowledge inconsistencies}.

Recent work has begun to study inconsistencies within Wikipedia. For example, WikiContradict constructs a benchmark of conflicting textual claims and evaluates whether LLMs can recognize incompatible answers~\cite{hou-etal-2024-wikicontradict}. WikiCollide/CLAIRE detects corpus-level inconsistencies by surfacing conflicting Wikipedia passages for human review~\cite{semnani-etal-2025-detecting}. These studies show that inconsistencies are common and that automatic systems can help finding them. However, they mostly focus on textual evidence which leaves an important gap: how to systematically detect and categorize inconsistencies across text, tables, and KGs.

To address this gap, we need a setting where the same information needs can be grounded in multiple modalities. Table-based question answering provides such a hinge point. Its questions are grounded in Wikipedia tables or tables and table-relevant text, while recent Text-to-SPARQL models allow the same questions to be queried against Wikidata. This creates a direct comparison: the table-based answer represents Wikipedia table-text evidence, and the SPARQL result retrieves KG evidence. The consistent agreement provides mutual support. Inconsistent disagreement provides a signal for possible knowledge conflicts, missing KG structure, or temporal drift.

In this paper, we introduce the task of \emph{modality-level inconsistency detection}: given a table-based question, its Wikipedia-grounded answer, and a KG answer obtained through Text-to-SPARQL querying, the goal is to determine whether the modalities agree and to characterize the mismatch when they do not.

\begin{figure}[htbp]
    \centering
    \includegraphics[
        width=1.0\linewidth,
        height=0.45\linewidth,
        keepaspectratio=false
    ]{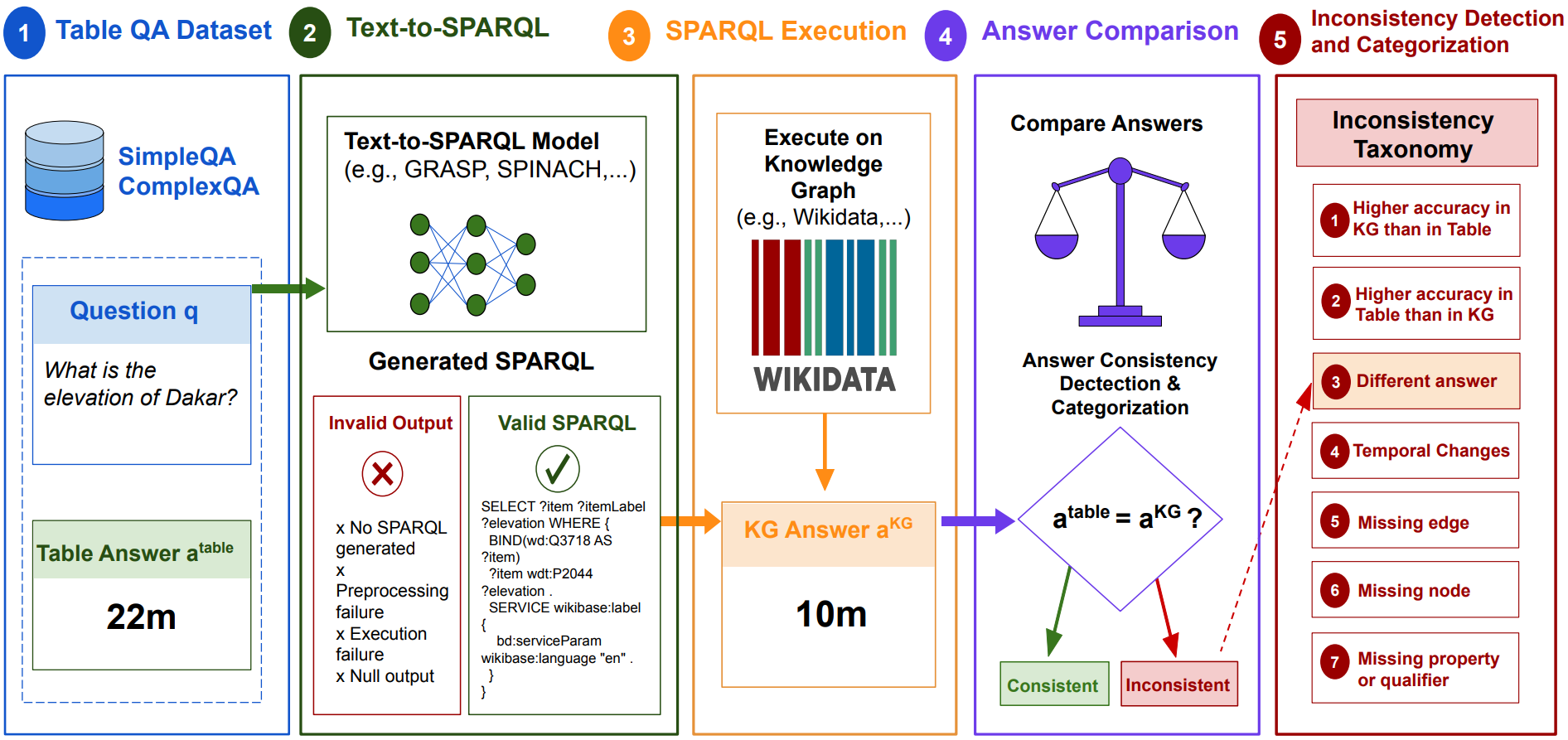}
      \caption{Workflow for detecting and categorizing modality-level inconsistencies by comparing table and KG answers using a knowledge inconsistency taxonomy.}
    \label{fig:teaser-figure}
\end{figure}

We introduce \textsc{Kontrast} (\textbf{K}nowledge \textbf{Contrast}), an automatic system for detecting and categorizing inconsistencies across text, tables, and KGs. Given a table-based QA instance, \textsc{Kontrast} retrieves Wikidata KG evidence with Text-to-SPARQL and compares it with the Wikipedia-table-grounded answer using rule-based matching and LLM-based reasoning. Its central idea is to treat cross-modal disagreement not as noise, but as a signal for knowledge reconciliation. Figure~\ref{fig:teaser-figure} depicts the overview of our system. Our contributions are:
\begin{itemize}
    \item We introduce modality-level inconsistency detection across text, tables, and KGs.
    \item We propose a taxonomy of cross-modal knowledge inconsistencies, covering direct conflicts, granularity differences, temporal changes, and KG incompleteness.
    \item We present \textsc{Kontrast}, a system for detecting and categorizing inconsistencies across text, tables, and KGs.
    \item We generate a new table-based QA dataset suitable for evaluating \textsc{Kontrast} and we show that cross-modal inconsistencies are measurable at scale and can reveal where modalities complement or correct each other.
\end{itemize}

\section{Preliminaries and Related Work}
\label{sec:related-work}

\subsection{Knowledge Conflict}
Knowledge conflict can be divided into three main directions:
(i) \textit{Uncertainty in KG construction} studies uncertainty arising from extracting, aligning, and fusing heterogeneous data into a KG. These discrepancies, termed knowledge deltas, reflect the semantic nature of knowledge conflicts such as invalidity, ambiguity, vagueness, fuzziness, timeliness, and incompleteness~\cite{jarnac-et-al-2025-uncertainty,djebri-etal-2019-label}. In parallel, WikiConflict offers a Wikidata revision-based dataset for conflicting data reconciliation in KG construction~\cite{jarnac-etal-2025-wikiconflict}.
(ii) \textit{Corpus-level inconsistency} studies contradictions inside large knowledge corpora such as Wikipedia. WikiContradict~\cite{hou-etal-2024-wikicontradict} and CLAIRE~\cite{semnani-etal-2025-detecting} show that real Wikipedia passages can support incompatible answers and that LLM-based systems can help automate surfacing such conflicts for human review. However, these works mainly treat conflict as disagreement among textual claims or retrieved passages.
(iii) \textit{LLM and RAG conflict} studies disagreement among retrieved sources or between evidence and a model's parametric knowledge. Prior work examines how contextual and parametric knowledge affect QA~\cite{longpre-etal-2021-entity}, and analyzes conflicts from outdated knowledge, ambiguity, misinformation, and inconsistent retrieval~\cite{wan-etal-2024-evidence,liu-etal-2025-conflicting,cattan-etal-2025-dragged,wang-etal-2025-conflicting}. Surveys distinguish context--memory, inter-context, and intra-memory conflicts~\cite{xu-etal-2024-knowledge-conflicts}, while conflict-aware QA benchmarks test whether systems can separate consensus from conflicting answers~\cite{nachshoni-etal-2025-consensus}.

\subsection{Table-based QA}
Tables store rich structured knowledge, and existing table-based QA datasets can again be grouped into three categories:
(i) \textit{Table QA} uses a single table as evidence, often through table-to-text generation or question annotation~\cite{nan-etal-2022-fetaqa,parikh-etal-2020-totto,pasupat-liang-2015-compositional-wititq,cheng-etal-2022-hitab}. Examples include FeTaQA that requires free-form reasoning over table cells, ToTTo that focuses on faithful generation from highlighted cells, and NQ-Table that extracts naturally asked questions grounded in Wikipedia tables or infoboxes~\cite{herzig-etal-2021-open,kwiatkowski-etal-2019-natural}.
(ii) \textit{Table-Text QA} combines tables with textual passages. HybridQA and OTT-QA require multi-hop reasoning across both modalities~\cite{chen-etal-2020-hybridqa,chen-etal-2021-ottqa}. QAMPARI supports multi-answer questions with distributed Wikipedia text paragraph evidence~\cite{amouyal-etal-2023-qampari}, while MoNaCo \cite{wolfson-etal-2025-monaco} targets time-consuming questions requiring reasoning chains over dozens of tables and passages. Meanwhile, SportReason focuses on sports-domain reasoning and reconstructs evidence from Table-Text QA datasets, providing a multi-table with multi-text scenario \cite{fu-etal-2025-sportreason}.
(iii) \textit{Heterogeneous QA} goes beyond one table and its passages. CompMix introduces compositional questions over KGs, web tables, and text~\cite{christmann-etal-2024-compmix}, while TANQ uses sentences, tables, and infoboxes to produce summary-table answers~\cite{feigenblat-etal-2025-tanq}. However, these benchmarks usually assume coherent evidence, leaving inconsistencies across texts, tables, and KGs largely under-explored. The dataset we propose is built from all these benchmarks.

\subsection{Text-to-SPARQL}
KBQA answers natural language questions by translating them into SPARQL queries~\cite{perez2006sparql} over knowledge bases such as Wikidata~\cite{vrandecic2014wikidata}. Earlier systems mainly rely on semantic parsing~\cite{yih-etal-2015-semantic,xu-etal-2023-fine}, often requiring supervised query annotations. Recent LLM-based methods move toward zero-shot and agentic settings. For instance, GRASP~\cite{walter-bast-2025-grasp} proposes a zero-shot SPARQL generation framework that combines LLM reasoning with dynamic graph exploration, achieving strong generalization across arbitrary RDF knowledge bases without fine-tuning. Similarly, SPINACH~\cite{liu-etal-2024-spinach} proposes a challenging real-world KBQA benchmark derived from complex user queries and an LLM-guided agent that mimics human expert SPARQL construction through iterative navigation. Both show promising results on multiple KBQA benchmarks~\cite{usbeck-etal-2017-qald7,usbeck-etal-2023-qald10,xu-etal-2023-fine}. Existing KBQA work focuses on query accuracy and answer retrieval. In contrast, we use Text-to-SPARQL as a bridge between table questions and KG evidence. This setting exposes discrepancies caused by incomplete, outdated, or conflicting knowledge, motivating our framework to detect and categorize inconsistencies across text, tables, and KGs.

\section{Taxonomy of Knowledge Inconsistency}
\label{sec:taxonomy}
In our setting, Wikipedia provides table and table-text evidence, while Wikidata provides KG evidence through entities, relations, properties, and qualifiers. Following prior work on knowledge deltas in KG construction, mismatches between these heterogeneous sources may arise from differences in granularity or from contradictions \cite{jarnac-et-al-2025-uncertainty}. Since different conflicts require different interpretations and actions \cite{cattan-etal-2025-dragged}, we classify answer-level mismatches into a taxonomy of conflicts across text, tables, and KGs.

We derived the taxonomy by inspecting the Wikipedia pages from which the table-based questions were constructed and manually comparing their answers with the corresponding Wikidata item pages. This process revealed recurring inconsistency patterns, which we refined by examining their causes and relating them to prior work~\cite{jarnac-et-al-2025-uncertainty}. Our proposed taxonomy focuses on answer-level inconsistencies across modalities, rather than on the intrinsic semantic nature of uncertainty.

Overall, our taxonomy distinguishes granularity mismatches from direct contradictions, temporal validity conflicts, and structural incompleteness in the KG, at the schema level (a property or a qualifier does not exist) or at the instance level (an entity or an edge is missing). This distinction is useful for downstream knowledge maintenance: granularity-related cases indicate which source should be enriched; different-answer and temporal conflicts require verification against context and time; and missing nodes, edges, properties, or qualifiers provide actionable signals for KG completion and schema refinement.

\begin{table*}[htbp]
\centering
\scriptsize
\setlength{\tabcolsep}{4pt}
\renewcommand{\arraystretch}{1.18}

\begin{tabularx}{\textwidth}{@{}
p{0.23\textwidth}
X
X
@{\hspace{\tabcolsep}\hspace{8pt}\hspace{\tabcolsep}}
p{0.20\textwidth}
@{}}
\toprule
\textbf{Label} & \textbf{Question} & \textbf{Table-based answer} & \textbf{KG Answer} \\
\specialrule{0.3pt}{0pt}{0pt}

Same answer &
Who was the king of England in 1756? &
George II &
George II of Great Britain \\
\midrule

Higher accuracy in KG than in Table &
In what movie did Ian Charleson play Eric Liddell, and what year did the movie come out? &
Ian Charleson played Eric Liddell in \emph{Chariots of Fire}, in 1981 &
\makecell[tl]{Chariots of Fire,\\
30 Mar 1981;\\
31 Mar 1981;\\
15 May 1981;\\
26 Sep 1981;\\
9 Apr 1982;\\
7 May 1982} \\
\midrule

Higher accuracy in Table than in KG &
Who starred in the \emph{pirates of the caribbean} &
Johnny Depp | Geoffrey Rush | Kevin McNally | Orlando Bloom | Keira Knightley | Jack Davenport | Jonathan Pryce &
Johnny Depp \\
\midrule

Different answer &
What is the elevation of Dakar? &
22 m &
10 m \\
\midrule

Temporal changes &
When is the season finale of designated survivor? &
May 16, 2018 &
7 June 2019 \\
\midrule

Missing edge &
Who directed (P57) the TV series \emph{Decoupled (Q110323803)} ? &
Hardik Mehta (Q35488479) &
-   \\
\midrule

Missing node &
What city is the university that taught Angie Barker located in ? &
Johnson City &
- \\
\midrule

Missing property/qualifier &
Who ran the fastest in the Men's 100 metres in the first semi-final of the 2012 Summer Olympics? &
Justin Gatlin (9.82), ahead of Churandy Martina (9.91) and Asafa Powell (9.94). &
Usain Bolt | 9.87 \\
\bottomrule
\end{tabularx}

\caption{Examples of inconsistencies between Wikipedia table-based answers and Wikidata KG answers. Questions come from existing benchmarks. The symbol ``-'' indicates that no SPARQL query can be generated and thus no result can be provided from the KG.}

\label{tab:taxonomy}
\end{table*}
\vspace{-0.5cm}
\noindent
\textbf{Same answer.} This category occurs when the Wikipedia table-based answer and the Wikidata KG answer refer to the same real-world entity, value, or fact at the same level of accuracy and coverage. These include differences in naming, aliases, language-tagged labels, or minor formatting. For example, the table-based answer \emph{George II} and the KG answer \emph{George II of Great Britain} refer to the same monarch. This category serves as a consistent baseline: the modalities agree, even if they express the answer at slightly different lexical or representational levels.

\noindent
\textbf{Higher accuracy in KG than in Table.} This inconsistency occurs when the KG answer has a higher level of precision or/and completeness than the table-based table-based answer. The two sources are not necessarily contradictory. Instead, they differ in granularity or specificity. For example, in Table~\ref{tab:taxonomy}, the table-based answer states that the movie \emph{Chariots of Fire} came out in 1981, while Wikidata contains multiple release dates, with a day-month-year granularity, depending on countries. The table-based answer derived from a table gives a valid answer, while Wikidata KG provides more detailed information.

\noindent
\textbf{Higher accuracy in Table than in KG.}
This inconsistency occurs when the table-based answer has a higher level of precision or/and completeness than the KG answer. In the cast example of Table~\ref{tab:taxonomy}, the table-based answer lists several actors who all exist in Wikidata, while the KG result returns only Johnny Depp. In this case, the KG answer is a subset of the table-based answer.

\noindent
\textbf{Different answer.}
This conflict occurs when the table-based answer and the KG answer provide incompatible values for the same question. Unlike specificity-related cases, the mismatch cannot be resolved by adding detail or choosing a more precise answer. For example, a table in Wikipedia reports Dakar's elevation as 22 m, while Wikidata reports 10 m in Table~\ref{tab:taxonomy}. In this case, the two sources may refer to elevations measured at different locations within Dakar.

\noindent
\textbf{Temporal changes.}
This conflict occurs when the correct answer depends on the temporal state of the source. In Table~\ref{tab:taxonomy}, the question \emph{When is the season finale of Designated Survivor?} gets the table-based answer of \emph{May 16, 2018}, corresponding to the Season 2 finale. This answer was true when this TV series was expected to end after two seasons.  Wikidata returns \emph{June 7, 2019}, corresponding to the Season 3 finale, which is valid today since the TV series was finally renewed and later concluded. Thus, these answers can be considered both valid depending on the temporal snapshot considered. This case is actually frequent, occurring for example with population census, rankings, geographic measurements, or event schedules which are all time dependent.

\noindent
\textbf{Missing edge.}
This inconsistency occurs when Wikidata contains the necessary entities and property but lacks the statement needed to connect them. In the \emph{Decoupled} example in Table~\ref{tab:taxonomy}, both the TV series \emph{Decoupled} (Q110323803)\footnote{\url{https://www.wikidata.org/wiki/Q110323803}} and the director Hardik Mehta (Q35488479)\footnote{\url{https://www.wikidata.org/wiki/Q35488479}} exist in Wikidata, but they are not linked with the \emph{director} property (P57).\footnote{\url{https://www.wikidata.org/wiki/Property:P57}} Thus, the KG contains the required entities and schema but lacks the statement needed to derive the table-based answer. This category captures KG incompleteness at the instance level.

\noindent
\textbf{Missing node.}
This inconsistency occurs when the KG lacks an entity needed to represent the answer or an intermediate reasoning step. In Table~\ref{tab:taxonomy}, the question \emph{What city is the university that taught Angie Barker located in?} comes from the Southern Conference Hall of Fame Wikipedia page.\footnote{\url{https://en.wikipedia.org/wiki/Southern_Conference_Hall_of_Fame}} While most players on the page are linked to Wikipedia entries, Angie Barker has no corresponding Wikipedia or Wikidata entity. This is another form of incompleteness in the KG at the instance level which lacks the entity/node needed to answer the question.

\noindent
\textbf{Missing property or qualifier.}
This inconsistency occurs when the KG contains the relevant entities, but lacks the property or the qualifier needed to express the answer at the required precision. In the Olympic semifinal example in Table~\ref{tab:taxonomy}, the question asks for the fastest runner in the first semifinal, whereas the KG answer returns Usain Bolt who is the winner of the final. Although the Wikidata page\footnote{\url{https://www.wikidata.org/wiki/Q734020}} includes the \emph{stage reached} property (P2443) with value \emph{semi-final} (Q599999), it does not distinguish the first semifinal from the second one. Thus, the KG schema lacks the fine-grained qualifier needed to represent the answer which is an incompleteness at the schema level of the KG.

\section{\textsc{Kontrast}: A Framework for Detecting and Categorizing Inconsistencies Across Modalities}
\label{sec:approach}
We define \emph{modality-level inconsistency detection} as a task over question--answer pairs grounded in heterogeneous Wikimedia resources. Given a question $q$, Wikipedia provides a table-based answer $a^{\mathrm{table}}$ supported by table or table-text evidence, while Wikidata provides a KG answer $a^{\mathrm{KG}}$ derived from KG evidence. The goal is to determine whether the two answers express the same fact:
\[
a^{\mathrm{table}} \equiv a^{\mathrm{KG}}.
\]
If they are equivalent, the instance is labeled \textsc{Same}. Otherwise, it is assigned one of the inconsistency labels in Table~\ref{tab:taxonomy}. Unlike standard QA, where one correct answer is sufficient, our task asks whether two modalities agree on the same information need.

\noindent
Our approach can be decomposed into three steps:
\begin{enumerate}
    \item \textbf{KG evidence retrieval:} generate and execute a SPARQL query over the KG for each table-text-based question.
    
    \item \textbf{Inconsistency detection:} compare the KG answer with the table-based answer to decide whether they are semantically equivalent or not.
    
    \item \textbf{Inconsistency categorization:} assign labels describing the inconsistency, distinguishing differences in terms of granularity \cite{jarnac-et-al-2025-uncertainty}, contradictions \cite{hou-etal-2024-wikicontradict}, temporal mismatches, and KG incompleteness.
\end{enumerate}

\subsection{KG evidence retrieval}
\textbf{Table-Text-based QA input.}
First, we select table-text based QA datasets whose answers are grounded in Wikipedia text and table evidence and can be verified against KG. Each instance contains a question $q$ and a table-based answer $a^{\mathrm{table}}$, supported by a Wikipedia table, infobox, or table with related text~\cite{Wei:TRIPLET2026,ettaleb:hal-05643904}. The question defines the information need, and the table-based answer represents the answer derived from Wikipedia tables and text.

\textbf{Text-to-SPARQL translation.}
A Text-to-SPARQL model translates a natural language question $q$ into a SPARQL query over a KG. This step maps natural language to symbolic KG operations over entities, properties, and qualifiers. We consider two recent systems: GRASP and SPINACH. GRASP is a zero-shot SPARQL generation framework that searches for relevant IRIs, executes intermediate queries, and composes a final executable query. SPINACH is an in-context KBQA agent that simulates expert query construction by navigating entities and properties over multiple steps.

We conducted a pilot study and we observed that GRASP produced more executable queries on average than SPINACH. Its output usually contains a self-contained SPARQL query encoding the full reasoning path, which makes execution and comparison straightforward. By contrast, SPINACH often encodes only the final reasoning step in SPARQL while relying on intermediate exploration outside the final query, making the result harder to interpret. GRASP is also more suitable for scalable experiments. While SPINACH is primarily configured for OpenAI models, GRASP supports OpenAI-compatible endpoints and locally served open-source models such as Qwen~\cite{yang2025qwen3}. We therefore use GRASP as the Text-to-SPARQL component in our experiments.

\subsection{Inconsistency detection}
\textbf{SPARQL execution and normalization.}
We separate the outputs generated into \emph{valid} and \emph{invalid} cases. A case is \emph{valid} if the model produces an executable SPARQL query, regardless of whether the query returns a non-empty set or not. For each valid query, we execute it over Wikidata and collect the returned values as $a^{\mathrm{KG}}$. Value-bearing results are passed to comparison and categorization. Empty result sets are stored separately. Invalid cases include null outputs, missing queries, execution failures, and preprocessing failures. We retain them because they may reflect not only Text-to-SPARQL errors, but also gaps in KG coverage or schema expressiveness. Before comparison, we normalize the raw KG output by removing table metadata, Wikidata identifiers, and datatype annotations, yielding a compact textual representation of $a^{\mathrm{KG}}$.

\textbf{Answer comparison.}
We compare $a^{\mathrm{table}}$ and $a^{\mathrm{KG}}$ using semantic equivalence rather than exact string matching. An instance is labeled \textsc{Same} if the two answers denote the same entity, value, or fact despite surface form variation, such as ``George II'' and ``George II of Great Britain''. Otherwise, it is treated as an inconsistency candidate. For example, ``22 m'' and ``10 m'' for Dakar's elevation are conflicting values and are passed to categorization.

\subsection{Inconsistency categorization}
\label{sec:inconsistency-categorization}
\textbf{Simple heuristics.}
Categorization begins with lightweight automatic heuristics. We normalize $a^{\mathrm{table}}$ and $a^{\mathrm{KG}}$ into sets of answer units, with special handling for dates and multi-value answers. Each table--KG answer pair receives an alignment score based on exact normalized matching, date-aware matching, or SBERT similarity~\cite{reimers-2019-sentence-bert} for aliases and minor lexical variation.

We compute recall from table answers to KG answers and precision from KG answers to table answers. An instance is labeled \textsc{Same} only when the two sets have the same size and both precision and recall are at least 0.95. This threshold was chosen through iterative human inspection of borderline cases, preserving clear matches while separating cases with missing or extra answers.
If the KG covers the table answer but adds more specific values, we label it \textsc{Higher accuracy in KG than in Table}. If the table answer contains additional values missing from the KG, we label it \textsc{Higher accuracy in Table than in KG}. Cases not confidently assigned by these rules are deferred to a judge LLM.

\textbf{LLM-as-a-judge.}
The remaining ambiguous cases are categorized with an LLM-as-a-judge protocol. The prompt contains five label definitions and three in-context examples per label, forming a 15-shot prompt. Each input includes the question, the table answer $a^{\mathrm{table}}$, and the KG answer $a^{\mathrm{KG}}$. The model outputs the taxonomy label, severity level, and a concise explanation. This stage is designed to flag meaningful inconsistency labels for human review.

\section{Experiments and Results}
\label{sec:experiments}

\subsection{Data Collection}
We construct a table-text-based QA collection to evaluate whether questions grounded in Wikipedia tables, infoboxes, and related passages can be translated into executable SPARQL queries over Wikidata KG. Each QA pair serves as a compact representation of the underlying table or semi-structured evidence: the question specifies the information needed, and the table-based answer captures the answer derived from Wikipedia tables and text. Rather than reusing existing benchmarks directly, we re-organize selected instances into a unified Text-to-SPARQL evaluation collection. We select data according to three criteria:
\begin{itemize}
    \item \textbf{Grounded evidence:} questions must be grounded in Wikipedia tables, infoboxes, or tables with related textual passages.
    \item \textbf{Reasoning diversity:} the collection should cover a broad range of reasoning complexity and skills, including direct lookup, multi-hop reasoning, temporal reasoning, composition, aggregation, and numerical reasoning.
    \item \textbf{Comparable answers:} table-based answers must be comparable with Wikidata query results. We therefore prioritize short strings, keywords, entities, and entity lists over long free-form answers.
\end{itemize}
\noindent
Table~\ref{tab:data_sources} summarizes the source datasets that have been used. We group them into \textit{SimpleQA} and \textit{ComplexQA}, which differ in the complexity of the information needed and the amount of reasoning required.

\paragraph{SimpleQA} contains natural or lightly structured questions that can be answered with direct table, infobox, or table-related textual evidence. It combines NQ-Table, simple CompMix questions without temporal, aggregation, or conjunction cues, and QAMPARI multi-answer questions~\cite{herzig-etal-2021-open,kwiatkowski-etal-2019-natural,christmann-etal-2024-compmix,amouyal-etal-2023-qampari}.

\begin{table}[t]
\centering
\scriptsize
\setlength{\tabcolsep}{3pt}
\renewcommand{\arraystretch}{1.02}
\begin{tabularx}{\columnwidth}{@{}lXllr@{}}
\toprule
\textbf{Group} & \textbf{Source} & \textbf{Creator} & \textbf{Evidence Type} & \textbf{\# Questions} \\
\midrule

\multirow{3}{*}{SimpleQA}
& NQ-Table~\cite{herzig-etal-2021-open,kwiatkowski-etal-2019-natural}
& Human & Table or infobox & 966 \\
& CompMix-simple~\cite{christmann-etal-2024-compmix}
& Human & Table and infobox & 326 \\
& QAMPARI~\cite{amouyal-etal-2023-qampari}
& Template & Table and text & 78 \\
\cmidrule(lr){2-5}
& \textbf{Subtotal} & & & \textbf{1,370} \\

\midrule

\multirow{6}{*}{ComplexQA}
& CompMix-infobox~\cite{christmann-etal-2024-compmix}
& Human & Infobox & 300 \\
& CompMix-table~\cite{christmann-etal-2024-compmix}
& Human & Table & 300 \\
& MoNaCo-time~\cite{wolfson-etal-2025-monaco}
& Human & Table and text & 150 \\
& MoNaCo~\cite{wolfson-etal-2025-monaco}
& Human & Table and text & 150 \\
& OTT-QA~\cite{chen-etal-2021-ottqa}
& Template & Table and text & 400 \\
& SportReason~\cite{fu-etal-2025-sportreason}
& LLM-RAG & Table, text, and infobox & 200 \\
\cmidrule(lr){2-5}
& \textbf{Subtotal} & & & \textbf{1,500} \\

\midrule
\textbf{Total} & & & & \textbf{2,870} \\
\bottomrule
\end{tabularx}
\caption{Dataset sources for Text-to-SPARQL generation.}
\label{tab:data_sources}
\end{table}

\paragraph{ComplexQA} contains questions that require richer evidence integration or more complex reasoning. It includes the longest table- and infobox-grounded questions from CompMix, time-dependent and non-time-dependent questions from MoNaCo, decontextualized OTT-QA development questions, and SportReason questions for numerical reasoning over text, tables, and infoboxes~\cite{christmann-etal-2024-compmix,wolfson-etal-2025-monaco,chen-etal-2021-ottqa,fu-etal-2025-sportreason,feigenblat-etal-2025-tanq}.

\subsection{Experiment Setup}
We evaluate \textsc{Kontrast} with three Qwen3 backbones: Qwen3-235B-A22B-Thinking, Qwen3-30B-A3B-Thinking, and Qwen3-4B-Instruct~\cite{yang2025qwen3}. These models cover different capabilities and inference settings, allowing us to test how model size (4B, 30B and 235B) affects modality-level inconsistency detection. We use GRASP for Text-to-SPARQL generation~\cite{walter-bast-2025-grasp}.

For each instance, we provide only the question text to the Text-to-SPARQL model. The generated query is executed against Wikidata, and the resulting KG answer is compared with the table-based answer from the original table-QA datasets.

All SPARQL queries are executed on the QLever Wikidata endpoint,\footnote{\url{https://qlever.dev/api/wikidata}} using the fixed 2025-05-10 Wikidata dump released with GRASP.\footnote{\url{https://ad-publications.cs.uni-freiburg.de/grasp/kg-index-search-index/}} This snapshot makes results reproducible and avoids changes from live Wikidata updates.

For categorization, we first apply rule-based heuristics with SBERT matching using \texttt{all-MiniLM-L6-v2}. Remaining cases are classified by a Qwen3-30B-A3B-Thinking judge~\cite{yang2025qwen3} with a 15-shot taxonomy prompt{}) (Section~\ref{sec:inconsistency-categorization}).

For taxonomy analysis, we use the Analysis Set in Table~\ref{tab:sparql_qa_execution_summary}, excluding outputs with more than 10 rows, because the Text-to-SPARQL model serializes only the first five and last five rows, and duplicate questions from source datasets. This ``row'' filter removes 251 cases for Qwen3-4B-Instruct (8.72\%), 97 for Qwen3-30B-Thinking (3.37\%), and 95 for Qwen3-235B-Thinking (3.30\%). Duplicate removal affects 12, 14, and 17 value-bearing cases, respectively.

\subsection{Results}
\paragraph{SPARQL validity.}
We first measure whether generated SPARQL queries execute successfully. As shown in Table~\ref{tab:sparql_qa_execution_summary}, Qwen3-30B-Thinking achieves the highest validity, with 2,576 valid outputs out of 2,870 cases (89.8\%). It performs best on both SimpleQA (93.3\%) and ComplexQA (86.5\%). Qwen3-235B-Thinking ranks second overall (79.5\%), while Qwen3-4B-Instruct is lowest (71.1\%), especially on complex datasets.

\paragraph{From executable queries to usable answers.}
A valid query may still return an empty answer. Therefore, Table~\ref{tab:sparql_qa_execution_summary} separates valid outputs into empty and value-bearing results. Qwen3-30B-Thinking produces the biggest number of executable queries, but the lowest value-bearing rate among valid outputs (49.6\%). By comparison, Qwen3-235B-Thinking achieves the highest value-bearing rate among valid outputs (63.7\%). Thus, execution validity captures whether a query runs, while the value-bearing rate better indicates whether it yields usable KG evidence.

\begin{table*}[t]
\centering
\scriptsize
\setlength{\tabcolsep}{3pt}
\renewcommand{\arraystretch}{1.08}
\resizebox{1.0\textwidth}{!}{%
\begin{tabular}{@{}l rrr rrr rrr@{}}
\toprule
\textbf{Dataset}
& \multicolumn{3}{c}{\textbf{Qwen3-4B}}
& \multicolumn{3}{c}{\textbf{Qwen3-30B}}
& \multicolumn{3}{c}{\textbf{Qwen3-235B}} \\
\cmidrule(lr){2-4}
\cmidrule(lr){5-7}
\cmidrule(lr){8-10}
& \textbf{$n$} & \textbf{Exec.} & \textbf{Ans.}
& \textbf{$n$} & \textbf{Exec.} & \textbf{Ans.}
& \textbf{$n$} & \textbf{Exec.} & \textbf{Ans.} \\
\midrule

NQ-Table
& 966 & 79.6 & 56.8
& 966 & 94.7 & 54.5
& 966 & 91.2 & 67.7 \\

CompMix-simple
& 326 & 88.3 & 68.8
& 326 & 90.2 & 57.1
& 326 & 89.0 & 69.0 \\

QAMPARI
& 78 & 53.8 & 59.5
& 78 & 88.5 & 47.8
& 78 & 96.2 & 72.0 \\

\cmidrule(lr){1-10}
SimpleQA
& 1,370 & 80.2 & 60.1
& 1,370 & \textbf{93.3} & 54.8
& 1,370 & 90.9 & \textbf{68.2} \\

\midrule

CompMix-infobox
& 300 & 91.3 & 76.6
& 300 & 96.3 & 77.2
& 300 & 67.7 & 84.2 \\

CompMix-table
& 300 & 84.7 & 47.2
& 300 & 92.7 & 49.6
& 300 & 89.0 & 66.7 \\

MoNaCo-time
& 150 & 70.0 & 44.8
& 150 & 78.0 & 30.8
& 150 & 67.3 & 52.5 \\

MoNaCo
& 150 & 70.7 & 47.2
& 150 & 82.0 & 30.9
& 150 & 34.0 & 56.9 \\

OTT-QA
& 400 & 35.5 & 37.3
& 400 & 84.2 & 24.6
& 400 & 66.5 & 34.6 \\

SportReason
& 200 & 30.5 & 41.0
& 200 & 77.0 & 39.0
& 200 & 73.5 & 54.4 \\

\cmidrule(lr){1-10}
ComplexQA
& 1,500 & 62.8 & 53.6
& 1,500 & \textbf{86.5} & 44.5
& 1,500 & 69.0 & \textbf{58.3} \\

\midrule

Total
& 2,870 & 71.1 & 57.1
& 2,870 & \textbf{89.8} & 49.6
& 2,870 & 79.5 & \textbf{63.7} \\

\midrule

Analysis Set
& 1,778 & 100.0 & 50.7
& 2,465 & 100.0 & 47.3
& 2,169 & 100.0 & \textbf{61.8} \\

\bottomrule
\end{tabular}
}
\caption{Text-to-SPARQL execution quality by dataset and model. $n$ denotes the number of evaluated cases; for \textbf{Analysis Set}, it denotes the number of question--answer pairs. \textbf{Exec.} is the valid-execution rate, and \textbf{Ans.} is the value-bearing rate among valid executions. Bold marks the best result across models for summary rows.}
\label{tab:sparql_qa_execution_summary}
\end{table*}

\paragraph{Inconsistency patterns.}
Table~\ref{tab:taxonomy_summary_main} summarizes modality-level inconsistencies in the filtered Analysis Set. Across SimpleQA, ComplexQA, and all models, \textit{Different answer} is the largest category, covering (i) true cross-modal knowledge conflicts; (ii) Text-to-SPARQL translation noise where the model selects an incorrect entity or property; and (iii) missing property or qualifier cases (schema level), where the Text-to-SPARQL model falls back to the closest available KG entities. Granularity mismatches are also frequent. Across models, \textit{Higher accuracy in KG than in Table} accounts for 11.3--13.5\% of cases, while \textit{Higher accuracy in Table than in KG} accounts for 7.3--9.4\%. This shows that neither modality is uniformly more complete. Instead, tables, text, and KGs provide complementary evidence for knowledge reconciliation.

\begin{table}[t]
\centering
\scriptsize
\setlength{\tabcolsep}{3pt}
\renewcommand{\arraystretch}{1.03}
\resizebox{\columnwidth}{!}{%
\begin{tabular}{@{}lccc@{}}
\toprule
\textbf{Taxonomy Label}
& \textbf{SimpleQA}
& \textbf{ComplexQA}
& \textbf{All} \\
\cmidrule(lr){2-4}
& \multicolumn{3}{c}{\textbf{Qwen3-4B / Qwen3-30B / Qwen3-235B}} \\
\midrule

Same
& 29.8 / 35.9 / 38.4
& 34.0 / 36.6 / 38.9
& 31.6 / 36.2 / 38.6 \\

Higher accuracy in KG than in Table
& 10.8 / 11.0 / 12.4
& 17.1 / 12.3 / 9.7
& 13.5 / 11.6 / 11.3 \\

Higher accuracy in Table than in KG
& 9.7 / 10.3 / 9.9
& 4.2 / 6.7 / 8.7
& 7.3 / 8.7 / 9.4 \\

Different answer
& \textbf{47.4} / \textbf{38.6} / \textbf{36.9}
& \textbf{44.4} / \textbf{43.5} / \textbf{41.2}
& \textbf{46.1} / \textbf{40.9} / \textbf{38.7} \\

Temporal changes
& 2.3 / 4.1 / 2.4
& 0.3 / 0.9 / 1.4
& 1.4 / 2.7 / 2.0 \\

\cmidrule(lr){1-4}
Inconsistent rate
& 70.2 / 64.1 / \textbf{61.6}
& 66.0 / 63.4 / \textbf{61.1}
& 68.4 / 63.8 / \textbf{61.4} \\

\bottomrule
\end{tabular}
}
\caption{Distribution of modality-level inconsistency categories in the filtered Analysis Set. Each cell reports percentages in the model order shown in the header. Percentages are computed within each QA group and model. Bold marks the dominant non-\textit{Same} inconsistency category and the lowest inconsistent rate.}
\label{tab:taxonomy_summary_main}
\end{table}

\paragraph{Effect of model scale.} Larger models produce more reliable KG answers and fewer translation-induced mismatches. In the filtered Analysis Set, Qwen3-235B-Thinking has the lowest inconsistency rate among value-bearing cases (61.4\%), followed by Qwen3-30B-Thinking (63.8\%) and Qwen3-4B-Instruct (68.4\%). This trend holds for both SimpleQA and ComplexQA, where Qwen3-235B-Thinking also yields the highest proportion of \textit{Same} cases. Overall, stronger reasoning ability Text-to-SPARQL models improve the quality of downstream inconsistency analysis.

\paragraph{Interpreting question complexity.} The higher inconsistency rates on SimpleQA should not be interpreted as evidence that simple datasets contain more knowledge conflicts. Rather, SimpleQA questions are translated into SPARQL more reliably, making their KG answers more suitable for inconsistency analysis. ComplexQA questions require more difficult entity linking, relation selection, and compositional reasoning, which introduces additional Text-to-SPARQL noise. Therefore, group-level differences reflect both underlying cross-modal inconsistency and the current limitations of automatic SPARQL generation.

\subsection{Error Analysis}
\label{sec:error-analysis}
To analyze Text-to-SPARQL failures, we randomly inspect 15 invalid cases, selecting five from 3 sub categories: \textit{No SPARQL generated}, \textit{Execution failure}, and \textit{Preprocessing failure}. We exclude \textit{Null output}, since it leaves no generation trace to inspect. This inspection reveals four recurring patterns.

\paragraph{Empty results indicate missing edges or qualifiers.} In 50\% of verified errors, \texttt{empty\_sparql\_result} is caused by missing KG structural elements. For example, for the question \textit{Who is the publisher of Writers \& Lovers book?}, the Text-to-SPARQL model links the correct book entity, but Wikidata lacks the publisher edge \texttt{P123} statement with the answer. Similarly, for \textit{Did John Prine win a Grammy for Fair \& Square?}, Wikidata contains the relevant Grammy relation, but lacks the \texttt{for work} qualifier \texttt{P1686} needed to connect the award to the album.

\paragraph{No-query cases reflect ambiguous questions.} All inspected cases are due to questions that are under-specified without additional context. For example, \textit{Who is their current manufacturer of their uniform kits?} cannot be grounded reliably because the entity referred to by ``their'' is not available from the question alone.

\paragraph{Execution failures mix timeouts, hallucination, and missing entities.} Among the inspected cases, 40\% are endpoint temporary timeouts that succeed after re-running. Another 20\% are LLM hallucination, where the Text-to-SPARQL model produces an answer from parametric knowledge rather than executable KG evidence. The remaining 40\% are due to missing entities, such as \textit{Who played the character of Sevika in Arcane?}, where the character Sevika from \emph{Arcane}\footnote{\url{https://en.wikipedia.org/wiki/Arcane_(TV_series)}} is not available as a required Wikidata entity.

\paragraph{Preprocessing failures are malformed SPARQL.}
All inspected cases are caused by invalid SPARQL syntax. These are generation-quality errors rather than KG incompleteness. For example, some generated queries miss the predicate, producing an incomplete subject--predicate--object triple pattern.

\subsection{Human Evaluation}
\label{sec:human-evaluation}
We conduct a human evaluation to check the reliability of the two automatic labeling stages. The evaluation focuses on answer-level categorization: one annotator verifies whether the assigned taxonomy label correctly describes the relation between the table-based answer and the KG answer.

For the heuristic stage, we inspect 30 cases. For each of \textsc{Same}, \textsc{Higher Accuracy in Table than KG}, and \textsc{Higher Accuracy in KG than Table}, we sample the five highest-scoring and five lowest-scoring instances by alignment score, covering both clear and borderline cases. All 30 labels are judged correctly.

For the LLM-as-a-judge stage, we inspect 50 cases: the first 10 instances for each of its five output labels, \textsc{Same}, \textsc{Higher Accuracy in Table than KG}, \textsc{Higher Accuracy in KG than Table}, \textsc{Different Answer}, and \textsc{Temporal Changes}. All 50 labels are judged correctly revealing a perfect alignment between human evaluation and the LLM judge.

These results provide evidence that both stages produce reliable taxonomy labels on the inspected samples. They are intended as a quality check of the labeling pipeline, rather than a full annotation study.

\subsection{Limitations}

\paragraph{Question naturalness.} Our Text-to-SPARQL pipeline is sensitive to question naturalness. Human-written questions usually yield higher SPARQL validity, while template- or answer-derived questions can be harder to parse. For example, OTT-QA asks: \emph{What is the capacity of the mosque that is on the list of largest mosques, and that was opened to the public 22 February 1978?}. Such unnatural constraints make entity and property selection harder. Thus, a low validity may reflect a weakness regarding the formulation of the question rather than a weakness of the Text-to-SPARQL model or a KG incompleteness.

\paragraph{Limits of automatic structural diagnosis.} Our taxonomy includes three structural KG incompleteness: \textit{Missing edge}, \textit{Missing node}, and \textit{Missing property or qualifier}. These categories are difficult to detect automatically from SPARQL execution results alone. They can appear in both valid and invalid generations, and their surface forms often overlap with ordinary Text-to-SPARQL errors.

\paragraph{Missing property or qualifier.} A missing property or qualifier may still produce a valid query if the model falls back to a related KG fact. For example, for the question about the most passing yards in a single NFL game, the table-based answer is 554 yards,\footnote{\url{https://en.wikipedia.org/wiki/List_of_500-yard_passing_games_in_the_NFL}}, whereas the KG answer is Norm Van Brocklin, the player associated with that record. The query is executable, but the mismatch arises because Wikidata does not encode the record using \emph{record held} (P1000).\footnote{\url{https://www.wikidata.org/wiki/Property:P1000}} In invalid cases, the model may instead state that the required fact is not available in the KG. However, this cannot be accepted directly as evidence of KG incompleteness, because the failure may also come from selecting the wrong entity or property.

\paragraph{Missing edges and missing nodes.} The same ambiguity holds for missing edges and missing nodes. A missing edge may lead to an empty valid result, a fallback to a semantically close property, or an invalid generation claiming that the relation is unavailable. A missing node may cause the model to select a related but incorrect entity, or to report that no matching entity exists. In all cases, automatic execution signals are insufficient to distinguish true KG incompleteness from semantic parsing errors.

\section{Conclusion and Future Work}
\label{sec:conclusion}
We introduced modality-level inconsistency detection, a task for identifying and categorizing knowledge mismatches across text, tables, and KGs. We proposed a taxonomy of cross-modal inconsistency types and presented \textsc{Kontrast}, an automatic system that uses Text-to-SPARQL and LLM reasoning to detect and categorize cross-modal knowledge inconsistencies.

Our analysis shows that modality-level inconsistencies are measurable at scale. With Qwen3-235B-Thinking, 61.4\% of value-bearing cases are not fully aligned with the table answer. Excluding the mixed \textit{Different answer} category, 22.7\% of cases reflect interpretable differences, including Higher accuracy in KG than in Table cases, and Higher accuracy in Table than in KG cases, and temporal shifts. These cases provide actionable signals for correction, enrichment, and temporal verification across texts, tables and KGs.

These results show that text, tables, and KGs should not be treated as isolated knowledge sources. Their disagreements can reveal where one modality complements or corrects another. \textsc{Kontrast} provides a practical way to discover and classify cross-modal inconsistencies for human review, supporting knowledge correction and enrichment at scale. By formalizing the task, taxonomy, and evaluation setting, this work provides a basis for future research on cross-modal knowledge consistency.

Future work should extend human annotation to both value-bearing and invalid cases for the structural labels \textit{Missing edge}, \textit{Missing node}, and \textit{Missing property or qualifier}. This would separate KG incompleteness from Text-to-SPARQL errors and support automatic diagnosis. A further direction is reconciliation: deciding when to update the KG, refine table evidence, add qualifiers, or keep both answers under different temporal or contextual conditions.

\section*{Declaration of Use of Generative AI}
GPT-5.2 has been used to draft the workflow figure. All AI-generated content was reviewed and edited by the authors, who take full responsibility for the final manuscript.

\section*{Acknowledgments}
This work was partially supported by the French National Research Agency (Agence Nationale de la Recherche - ANR) under the ECLADATTA project, grant number ANR-22-CE23-0020.


\bibliographystyle{splncs04}
\bibliography{bibliography}

\end{document}